%% file: main.tex
\documentclass[orivec]{llncs}

\usepackage{latexsym,amssymb,amsmath}
\usepackage{url}
\usepackage[defblank]{paralist}
\usepackage{tikz,subfigure}
\usetikzlibrary{arrows,shapes,shapes.multipart,snakes,automata,backgrounds,petri,positioning,shadows,matrix,decorations.pathmorphing,fit,positioning,calc,backgrounds}
\usepackage{todonotes}
\usepackage{xspace}
\usepackage{pifont}
\usepackage{fixltx2e}
\usepackage{wrapfig}
\usepackage{listings}
\usepackage{xcolor}
\usepackage{times}

\input{macros-style}
\input{macros-common}
\input{macros}

\title{Petri Nets with Parameterised Data: \\ Modelling and Verification}
\subtitle{(Extended Version)}
\author{Silvio Ghilardi$^1$,  Alessandro Gianola$^{2,3}$, Marco Montali$^2$ \and Andrey Rivkin$^2$ }
\authorrunning{Ghilardi, Gianola, Montali, Rivkin}
\institute{%
$^1$Dipartimento di Matematica, Universit\`a degli Studi di Milano (Italy) \\
silvio.ghilardi@unimi.it\\
$^2$Faculty of Computer Science, Free University of Bozen-Bolzano (Italy)\\
\{gianola, montali, rivkin\}@inf.unibz.it \\
$^3$CSE Department, University of California San Diego (USA)\\
agianola@eng.ucsd.edu
}

\begin{document}

\maketitle

\begin{abstract}
During the last decade, various approaches have been put forward to integrate business processes with different types of data. Each of such approaches reflects specific demands in the whole process-data integration spectrum. One particular important point is the capability of these approaches to flexibly accommodate processes with multiple cases that need to co-evolve. In this work, we introduce and study an extension of coloured Petri nets, called catalog-nets, providing two key features to capture this type of processes. On the one hand, net transitions are equipped with guards that simultaneously inspect the content of tokens and query facts stored in a read-only, persistent database. On the other hand, such transitions can inject data into tokens by extracting relevant values from the database or by generating genuinely fresh ones. We  systematically encode catalog-nets into one of the reference frameworks for the (parameterised) verification of data and processes. We show that fresh-value injection is a particularly complex feature to handle, and discuss strategies to tame it. Finally, we discuss how catalog nets relate to well-known formalisms in this area.
\end{abstract}


\input{1-introduction}

\input{2-reponet}

\input{4-translation}

\input{verification}

\input{3-comparison}

\input{5-conclusions}

\bibliographystyle{splncs04}
\bibliography{mybib}

\newpage
\appendix

\input{0-appendix}

\input{1-appendix}

\end{document}

%% file: macros-style.tex
\makeatletter
\g@addto@macro\normalsize{%
\setlength{\abovecaptionskip}{0pt}
\setlength{\belowcaptionskip}{-10pt}
\setlength\abovedisplayskip{3pt}
\setlength\belowdisplayskip{3pt}
\setlength\abovedisplayshortskip{3pt}
\setlength\belowdisplayshortskip{3pt}
}
\makeatother

%% file: macros-common.tex
\newcommand{\R}{\ensuremath{\mathcal{R}}}

\newcommand{\set}[1]{\{#1\}}                      

\newcommand{\tup}[1]{\langle #1\rangle}            

\makeatletter
\g@addto@macro\normalsize{%
\setlength{\abovecaptionskip}{-2pt}
\setlength{\belowcaptionskip}{12pt}
\setlength\abovedisplayskip{3pt}
\setlength\belowdisplayskip{3pt}
\setlength\abovedisplayshortskip{3pt}
\setlength\belowdisplayshortskip{3pt}
}
\makeatother

\newcounter{dummy} 
\newcounter{dummy1} 
\newcounter{dummy2}
\newcounter{dummy3} 
\newcounter{dummy4}
\newcounter{dummy5} 
\newcounter{dummy6}

\usepackage[thmmarks,amsmath]{ntheorem}
\theorempreskip{1pt}
\theorempostskip{1pt}

\theoremseparator{.}
\theorembodyfont{\itshape}
\theoremsymbol{$\triangleleft$}
\newtheorem{theorem}[dummy]{Theorem}
\newtheorem{lemma}[dummy1]{Lemma}
\newtheorem{definition}[dummy2]{Definition}
\newtheorem{proposition}[dummy3]{Proposition}

\theorembodyfont{\normalfont}
\newtheorem{example}[dummy4]{Example}
\newtheorem{remark}[dummy5]{Remark}

\theoremstyle{nonumberplain}
\theoremheaderfont{\itshape}
\theorembodyfont{\normalfont}

\theoremseparator{.}
\theoremsymbol{\ensuremath{\dashv}}
\newtheorem{proof}[dummy6]{Proof}

\qedsymbol{\ensuremath{\dashv}}


%% file: macros.tex
\newcommand{\dbnet}{DB-net\xspace}
\newcommand{\dbnets}{DB-nets\xspace}
\newcommand{\clognet}{CLog-net\xspace}
\newcommand{\clognets}{CLog-nets\xspace}

\newcommand{\nucpn}{\ensuremath{\nu}-CPN\xspace}
\newcommand{\nucpns}{\ensuremath{\nu}-CPNs\xspace}

\newcommand{\mult}[1]{#1^\oplus}
\newcommand{\supp}[1]{\mathit{supp}(#1)}

\newcommand{\funsym}[1]{\mathtt{#1}}
\newcommand{\setsym}[1]{\mathit{#1}}
\newcommand{\restr}[2]{{
  \left.\kern-\nulldelimiterspace 
  #1 
  \vphantom{\big|} 
  \right|_{#2} 
  }}
\newcommand{\powerset}{\raisebox{.15\baselineskip}{\Large\ensuremath{\wp}}}

\newcommand{\schema}{\R}
\newcommand{\tfun}{\funsym{type}}

\newcommand{\free}[1]{\setsym{Free}(#1)}

\newcommand{\subst}{\theta}

\newcommand{\relname}[1]{\ensuremath{\mathit{#1}}} 
\newcommand{\cname}[1]{\ensuremath{\mathtt{#1}}\xspace} 

\newcommand{\typename}[1]{\mathbf{#1}}
\newcommand{\types}{\mathfrak{D}}
\newcommand{\type}{\mathcal{D}}

\newcommand{\dom}{\Delta}
\newcommand{\sigp}{\Gamma}

\newcommand{\naturals}{\mathbb{N}}

\newcommand{\vartype}{\funsym{type}}

\newcommand{\ans}{\ensuremath{\mathit{ans}}\xspace}

\newcommand{\cat}{\mathit{Cat}}
\newcommand{\val}{\mathit{Val}}

\newcommand{\cqd}{\mathtt{CQ_\types^{\neg}}\xspace}
\newcommand{\ucqd}{\mathtt{UCQ_\types^{\neg}}\xspace}

\newcommand{\qent}[3]{#1,#2\models #3}

\newcommand{\proj}[2]{\ensuremath{\relname{#1}.#2}}

\newcommand{\PK}[1]{\textsc{pk}(\relname{#1})}
\newcommand{\FK}[4]{\proj{#1}{#2}\rightarrow\proj{#3}{#4}}

\newcommand{\cval}[1]{\mathsf{#1}}

\newcommand{\conds}{\mathcal{C}_\types}

\newcommand{\varsin}[1]{\setsym{Vars}(#1)}
\newcommand{\constin}[1]{\setsym{Const}(#1)}

\newcommand{\inflow}{F_{in}}
\newcommand{\outflow}{F_{out}}

\newcommand{\transition}[1]{{\fontfamily{phv}\selectfont {#1}}}

\newcommand{\places}{P}
\newcommand{\cplaces}{P_c}

\newcommand{\coloring}{\funsym{color}}
\newcommand{\guass}{\funsym{guard}}
\newcommand{\nuvarset}{\Upsilon_{\types}}
\newcommand{\vars}{\mathcal{X}_{\types}}

\newcommand{\transitions}{T}

\newcommand{\tuples}[1]{\Omega_{#1}}

\newcommand{\invars}[1]{\setsym{InVars}(#1)}
\newcommand{\outvars}[1]{\setsym{OutVars}(#1)}

\newcommand{\net}{\mathcal{N}}
\newcommand{\varset}{\mathcal{V}_{\types}}

\newcommand{\nupns}{\ensuremath{\nu}-PNs\xspace}

\newcommand{\marked}[2]{\tup{#1,#2}}
\newcommand{\markedcat}[3]{\tup{#1,#2,#3}}

\newcommand{\enabledcat}[3]{#1[#2\rangle_{#3}}

\newcommand{\firecat}[4]{\enabledcat{#1}{#2}{#4}#3}

\newcommand{\tsys}[1]{\Lambda_{#1}}

\newcommand{\trans}{\rightarrow}

\newcommand{\safe}{\texttt{SAFE}\xspace}
\newcommand{\unsafe}{\texttt{UNSAFE}\xspace}
\newcommand{\mcmt}{\textsc{mcmt}\xspace}

 \definecolor{dbcolor}{HTML}{F39C12}
\definecolor{datacolor}{HTML}{FAD7A0}
\definecolor{activedatacolor}{HTML}{85C1E9}
\definecolor{viewcolor}{HTML}{5499C7}
\definecolor{actioncolor}{HTML}{2471A3}
\definecolor{netcolor}{HTML}{D576AE}
\definecolor{mgreen}{rgb}{0.128,0.428,0}
\definecolor{burntorange}{rgb}{0.8, 0.33, 0.0}
\definecolor{camouflagegreen}{rgb}{0.47, 0.53, 0.42}
\definecolor{copperrose}{rgb}{0.6, 0.4, 0.4}

\tikzstyle{place}=[circle,thick,draw=black,fill=white,minimum size=7mm,font=\fontsize{9}{144}\selectfont]
\tikzstyle{transition}=[rectangle,thick,draw=black,fill=gray!20,minimum size=7mm]
\tikzstyle{enabledtransition}=[rectangle,very thick,draw=green!75,fill=green!20,minimum size=7mm]
 \tikzstyle{container}=[rectangle,rounded corners,very thick,draw=black!75,fill=black!20,minimum height=7mm,minimum width=14mm]
\tikzstyle{rplace}=[circle,ultra thick,draw=violet!75,fill=violet!20,minimum size=7mm]

\tikzstyle{erbox}=[draw, fill=gray!20, minimum width=7em, text width=6.0em, text centered,
  minimum height=3em,rounded corners,drop shadow]

\tikzstyle{placelem}=[draw,
    cloud,
    cloud puffs = 15,
    minimum width=.2cm,
    minimum height=.2cm,
    fill=white,
    thick]

\tikzstyle{netelem}=[draw,
    cloud,
    cloud puffs = 20,
    minimum width=1.8cm,
    minimum height=1.8cm,
    fill=blue!10,
    ultra thick]

\tikzstyle{relationselem}=[placelem,fill=pink!20]
\tikzstyle{noopelem}=[placelem,fill=orange!20]
\tikzstyle{enteredplace}=[place,fill=yellow!20]
\tikzstyle{boundplace}=[place,fill=yellow!30]
\tikzstyle{guardokplace}=[place,fill=yellow!40]
\tikzstyle{updatedplace}=[place,fill=yellow!50]
\tikzstyle{violplace}=[place,fill=red!10]
\tikzstyle{constrokplace}=[place,fill=green!10]
\tikzstyle{docommitplace}=[place,fill=green!30]
\tikzstyle{dorollbackplace}=[place,fill=red!30]
\tikzstyle{arc}=[-stealth',thick]
\tikzstyle{readarc}=[-,thick]

\makeatletter
\pgfarrowsdeclare{X}{X}
{
  \pgfutil@tempdima=0.3pt%
  \advance\pgfutil@tempdima by.25\pgflinewidth%
  \pgfutil@tempdimb=5.5\pgfutil@tempdima\advance\pgfutil@tempdimb by.5\pgflinewidth%
  \pgfarrowsleftextend{+-\pgfutil@tempdimb}
  \pgfarrowsrightextend{0pt}
}
{
  \pgfutil@tempdima=0.3pt%
  \advance\pgfutil@tempdima by.25\pgflinewidth%
  \pgfsetdash{}{+0pt}
  \pgfsetroundcap
  \pgfsetmiterjoin
  \pgfpathmoveto{\pgfqpoint{-5.5\pgfutil@tempdima}{-6\pgfutil@tempdima}}
  \pgfpathlineto{\pgfqpoint{5.5\pgfutil@tempdima}{6\pgfutil@tempdima}}
  \pgfpathmoveto{\pgfqpoint{-5.5\pgfutil@tempdima}{6\pgfutil@tempdima}}
  \pgfpathlineto{\pgfqpoint{5.5\pgfutil@tempdima}{-6\pgfutil@tempdima}}
  \pgfusepathqstroke 
}
\makeatother

\makeatletter
\pgfarrowsdeclare{x}{x}
{
  \pgfutil@tempdima=0.3pt%
  \advance\pgfutil@tempdima by.25\pgflinewidth%
  \pgfutil@tempdimb=5.5\pgfutil@tempdima\advance\pgfutil@tempdimb by.5\pgflinewidth%
  \pgfarrowsleftextend{+-\pgfutil@tempdimb}
  \pgfarrowsrightextend{0pt}
}
{
  \pgfutil@tempdima=0.3pt%
  \advance\pgfutil@tempdima by.25\pgflinewidth%
  \pgfsetdash{}{+0pt}
  \pgfsetroundcap
  \pgfsetmiterjoin
  \pgfpathmoveto{\pgfqpoint{-3.5\pgfutil@tempdima}{-4\pgfutil@tempdima}}
  \pgfpathlineto{\pgfqpoint{3.5\pgfutil@tempdima}{4\pgfutil@tempdima}}
  \pgfpathmoveto{\pgfqpoint{-3.5\pgfutil@tempdima}{4\pgfutil@tempdima}}
  \pgfpathlineto{\pgfqpoint{3.5\pgfutil@tempdima}{-4\pgfutil@tempdima}}
  \pgfusepathqstroke 
}
\makeatother

\tikzset{
state/.style={
       rectangle,
       fill=yellow!5,
       rounded corners,
       draw=black,  thick,
       minimum height=2em,
       minimum width=1.5cm,
       inner sep=1pt,
       text centered,
       }
}

\usepackage{anyfontsize}
\usepackage{t1enc}
\colorlet{light-gray}{gray!20}
\definecolor{bronze}{rgb}{0.8, 0.5, 0.2}

\lstnewenvironment{mc}
{\lstset{
   frameround=ffff,
    language=C,    
    numbers=none,
    breaklines=true,
    breakatwhitespace=true,
    mathescape=true,
    tabsize=3,
    keywordstyle=\color{black}\bfseries, 
    morekeywords={:smt,define,:global},
    basicstyle=\footnotesize\ttfamily\color{black},
}}{}


%% file: 1-introduction.tex
 \section{Introduction}
The integration of control flow and data has become one of the most prominently investigated topics in BPM \cite{Reic12}. Taking into account data when working with processes is crucial to properly understand which courses of execution are allowed \cite{DDG17}, to account for decisions \cite{BaHW17}, and to explicitly accommodate business policies and constraints \cite{Dum11}.
Hence, considering how a process manipulates underlying volatile and persistent data, and how such data influence the possible courses of execution within the process, is central to understand and improve how organisations, and their underlying information systems, operate throughout the entire BPM lifecycle: from modelling and verification \cite{Hull08,CaDM13} to enactment \cite{KuWR11,MPFW13} and mining \cite{Aalst19}.
Each of such approaches reflects specific demands in the whole process-data integration spectrum. One key point is the capability of these approaches to accommodate processes with multiple co-evolving case objects \cite{Fahland19,AKMA19}. Several modelling paradigms have adopted to tackle this and other important features: data-/artifact-centric approaches \cite{Hull08,CaDM13}, declarative languages based on temporal constraints \cite{AKMA19}, and imperative, Petri net-based notations \cite{MonR17,Fahland19,PWOB19}. 

With an interest in (formal) modelling and verification, in this paper we concentrate on the latter stream, taking advantage from the long-standing tradition of adopting Petri nets as the main backbone to formalise processes expressed in front-end notations such as BPMN, EPCs, and UML activity diagrams. In particular, we investigate for the first time the combination of two different, key requirements in the modelling and analysis of data-aware processes. On the one hand, we support the creation of fresh (case) objects during the execution of the process, and the ability to model their (co-)evolution using guards and updates. Examples of such objects are orders and their orderlines in an order-to-cash process.    
On the other hand, we handle read-only, persistent data that can be accessed and injected in the objects manipulated by the process. Examples of read-only data are the catalog of product types and the list of customers in an order-to-cash process. Importantly, read-only data have to be considered in a \emph{parameterised} way. This means that the overall process is expected to operate as desired in a robust way, irrespectively of the actual configuration of such data.

 While the first requirement is commonly tackled by the most recent and sophisticated approaches for integrating data within Petri nets \cite{MonR17,Fahland19,PWOB19}, the latter has been extensively investigated in the data-centric spectrum \cite{DeLV16,MSCS20}, but only recently ported to more conventional, imperative processes with the simplifying assumptions that the process control-flow is block-structured (and thus 1-bounded in the Petri net sense) \cite{BPM19,CGGMR19-techrep-dab}. 
 
 In this work, we reconcile these two themes in an extension of coloured Petri nets (CPNs) called \emph{catalog-nets} (\clognets). On the one hand, in \ \clognet transitions are equipped with guards that simultaneously inspect the content of tokens and query facts stored in a read-only, persistent database. On the other hand, such transitions can inject data into tokens by extracting relevant values from the database or by generating genuinely fresh ones. 
 We systematically encode \clognets into the most recent version of \mcmt\footnote{\url{http://users.mat.unimi.it/users/ghilardi/mcmt/}} \cite{lmcs}, one of the few model checkers natively supporting the (parameterised) verification of data and processes \cite{CGGMR18,CGGMR19,MSCS20}. We show that fresh-value injection is a particularly complex feature to handle, and discuss strategies to tame it. We then stress that, thanks to this encoding, a relevant fragment of the model can be readily verified using \mcmt, and that verification of the whole model is at reach with a minor implementation effort.
  Finally, we discuss how catalog nets provide a unifying approach for some of the most sophisticated formalisms in this area, highlighting differences and commonalities.

%% file: 2-reponet.tex
\newcommand{\datatype}[1]{\texttt{#1}}

\section{The \clognet Formal Model}
\label{sec:clognets}

Conceptually, a \clognet integrates two key components. The first is a read-only persistent data storage, called \emph{catalog}, to account for read-only, parameterised data. The second  is a variant of CPN, called \nucpn \cite{MonR19}, to model the process backbone. Places carry tuples of data objects and can be used to represent: 
\begin{inparaenum}[\it (i)]
\item states of (interrelated) case objects, 
\item read-write relations, 
\item read-only relations whose extension is fixed (and consequently not subject to parameterisation), 
\item resources. 
 \end{inparaenum}
As in \cite{MonR19,Fahland19,PWOB19}, the net employs $\nu$-variables (first studied in the context of \nupns~\cite{RVFE11}) to inject fresh data (such as object identifiers). A distinguishing feature of \clognets is that transitions can have guards that inspect and retrieve data objects from the read-only, catalog.

%
%




\smallskip\noindent
\textbf{Catalog.} 
We consider a \emph{type domain} $\types$ as a finite set of pairwise disjoint data types accounting for the different types of objects in the domain. Each type $\type \in \types$ comes with its own (possibly infinite) \emph{value domain} $\dom_\type$, and with an equality operator $=_\type$. When clear from the context we simplify the notation and use $=$ in place of $=_\type$.
$R(a_1:\type_1,\ldots,a_n:\type_n)$ is a \emph{$\types$-typed relation schema}, where $R$ is a relation name and $a_i:\type_i$ indicates the $i$-th attribute of $R$ together with its data type. When no ambiguity arises, we omit relation attributes and/or their data types. 
A \emph{$\types$-typed catalog (schema)} $\schema_\types$ is a finite set of $\types$-typed relation schemas.
A \emph{$\types$-typed catalog instance $\cat$ over $\schema_\types$} is a finite set of facts $R(\cname{o_1},\ldots,\cname{o_n})$, where 
$R\in \schema$ and $\cname{o_i}\in \dom_{\type_i}$, for $i \in \set{1,\ldots,n}$. 
%

We adopt two types of \emph{constraints} in the catalog relations. First, we assume  the first attribute of every relation $R\in\schema_\types$ to be its \emph{primary key}, denoted as $\PK{R}$. Also, a type of such attribute should be different from the types of other primary key attributes. Then, for any $R,S\in\schema_\types$, $\FK{R}{a}{S}{id}$ defines that the projection $\proj{R}{a}$ is a \emph{foreign key} referencing $\proj{S}{id}$, where $\PK{S}=id$, $\PK{R}\neq a$ and $\tfun(id)=\tfun(a)$.
While the given setting with constraints may seem a bit restrictive, it is the one adopted in the most sophisticated settings where parameterisation of read-only data is tackled \cite{DeLV16,MSCS20}.

\newcommand{\prodcat}{\relname{ProdCat}}
\newcommand{\producttype}{\datatype{ProdType}}
\newcommand{\compatible}{\relname{Comp}}
\newcommand{\compid}{\datatype{CId}}
\newcommand{\trucktype}{\datatype{TruckType}}

\begin{example}
\label{ex:catalog}
Consider a simple catalog of an order-to-delivery scenario, containing two relation schemas.
Relation schema $\prodcat(p:\producttype)$ indicates the product types (e.g., vegetables, furniture) available in the organisation catalogue of products.
Relation schema $\compatible(c:\compid,p:\producttype,t:\trucktype)$ captures the compatibility between products and truck types used to deliver orders; e.g. one may specify that vegetables are compatible only with types of trucks that have a refrigerator. 
\end{example}

\smallskip\noindent
\textbf{Catalog queries.} 
We fix a countably infinite set $\varset$ of typed variables with a \emph{variable typing function} $\vartype: \varset \rightarrow \types$. Such function can be easily extended to account for sets, tuples and multisets of variables as well as constants. 
As  query language we opt for the union of conjunctive queries with inequalities and atomic negations that can be specified in terms of first-order (FO) logic extended with types. This corresponds to widely investigated SQL select-project-join queries with filters, and unions thereof.

A \emph{conjunctive query (CQ) with atomic negation} $Q$ over $\schema_\types$ has the form \[Q::=\varphi\,|\,R(x_1,\ldots,x_n)\,|\,\neg R(x_1,\ldots,x_n)\,|\,Q_1 \land Q_2\,|\,\exists x.Q\]
where
\begin{inparaenum}[\it(i)]
\item $R(\type_1,\ldots,\type_n)\in\schema$, $x\in\varset$ and each $x_i$ is either a variable of type $\type_i$ or a constant from $\dom_{\type_i}$;
\item $\varphi::=S(y_1,\ldots,y_m)\,|\,\neg\varphi\,|\,\varphi\land\varphi\,|\,\top$ is a \emph{condition} s.t. $S\in\sigp_\type$ and $y_i$ is either a variable of type $\type$ or a constant from $\dom_{\type}$.
\end{inparaenum}
$\cqd$ denotes the set of all such conjunctive queries, and $\free{Q}$ the set of all free variables (i.e., those not occurring in the scope of quantifiers) of query $Q$. 
$\conds$ denotes the set of all possible conditions, $\varsin{Q}$ the set of all variables in $Q$, and $\constin{Q}$ the set of all constants in $Q$.
Finally, $\ucqd$ denotes the set off all \emph{unions of conjunctive queries} over $\schema_\types$. Each query $Q\in\ucqd$ has the form $Q=\bigwedge_{i=1}^n Q_i$, with $Q_i\in\cqd$. 
 
A \emph{substitution} for a set $X = \set{x_1,\ldots,x_n}$ of typed variables
is a function $\subst: X \rightarrow \dom_\types$, such that
$\subst(x) \in \dom_{\vartype(x)}$ for every $x \in X$. An empty substitution is denoted as $\tup{}$.  A \emph{substitution $\subst$ for a query $Q$}, denoted as $Q\subst$, is a substitution for variables in $\free{Q}$. 
%
An \emph{answer to a query $Q$} in a catalog instance $\cat$ is a set of substitutions $\ans(Q,\cat) = \{ \subst: \free{Q} \rightarrow \val(\cat) \mid
\qent{\cat}{\subst}{Q}\}$, where $\val(\cat)$ denotes the set of all constants occurring in $\cat$ and $\models$ denotes standard FO entailment. 
%

\begin{example}
\label{ex:queries}
Consider the catalog of Example~\ref{ex:catalog}. Query $\prodcat(p)$ retrieves the product types $p$ present in the catalog, whereas given a product type value $\cval{veg}$, query $\exists c.\compatible(c,\cval{veg},t)$ returns the truck types $t$ compatible with $\cval{veg}$.
\end{example}


\medskip
\noindent
\textbf{\clognets.} We first fix some standard notions related to \emph{multisets}. Given a set $A$,  the \emph{set of multisets} over $A$, written $\mult{A}$, is the set of mappings of the form $m:A\rightarrow \mathbb{N}$.
Given a multiset $S \in \mult{A}$ and an element $a \in A$, $S(a) \in \mathbb{N}$ denotes the number of times $a$ appears in $S$.  
We write $a^n \in S$ if $S(a) = n$.  The support of $S$ is the set of elements that appear in $S$ at least once: $\supp{S} = \set{a\in A\mid S(a) > 0}$. We also consider the usual operations on multisets. Given $S_1,S_2 \in \mult{A}$:
\begin{inparaenum}[\it (i)]
\item $S_1 \subseteq S_2$ (resp., $S_1 \subset S_2$) if $S_1(a) \leq S_2(a)$ (resp., $S_1(a) < S_2(a)$) for each $a \in A$;
\item $S_1 + S_2 = \set{a^n \mid a \in A \text{ and } n = S_1(a) + S_2(a)}$;
\item if $S_1 \subseteq S_2$, $S_2 - S_1 = \set{a^n \mid a \in A \text{ and } n = S_2(a) - S_1(a)}$;
\item given a number $k \in \mathbb{N}$, $k \cdot S_1 = \set{a^{kn} \mid a^n \in S_1}$;
\item $|m|=\sum_{a\in A}m(a)$.
\end{inparaenum}
A multiset over $A$ is called empty (denoted as $\mult\emptyset$) iff $\mult\emptyset(a)=0$ for every $a\in A$.

We now define \clognets, extending \nucpns \cite{MonR19} with the ability of querying a read-only catalog. As in CPNs,  each \clognet  place has a color type, which corresponds to a data type or to the cartesian product of multiple data types from $\types$. 
Tokens in places are referenced via \emph{inscriptions} -- tuples of variables and constants. We denote by $\Omega_A$ the set of all possible inscriptions over a set $A$ and, with slight abuse of notation, use $\varsin{\omega}$ (resp., $\constin{\omega}$) to denote the set of variables (resp., constants) of $\omega\in\Omega_A$.
To account for fresh external inputs, we employ the well-known mechanism of $\nu$-Petri nets~\cite{RVFE11} and introduce a countably infinite set $\nuvarset$ of $\types$-typed \emph{fresh variables}, where for every $\nu \in \nuvarset$, we have that $\dom_{\vartype(\nu)}$ is countably infinite (this provides an unlimited supply of fresh values). We fix a countably infinite set of $\types$-typed variable $\vars = \varset \uplus \nuvarset$ as the disjoint union of ``normal"  ($\varset$) and fresh ($\nuvarset$) variables. 

\begin{definition}
\label{def:clognet}
A $\types$-typed \clognet $\net$ over a catalog schema $\schema_\types$ is a tuple  
$(\types, \schema_\types, \places,\transitions,\inflow,\outflow,\coloring,\guass)$, where:
\begin{compactenum}
%
\item $\places$ and $\transitions$ are finite sets of places and transitions, s.t. $P\cap T=\emptyset$;
\item $\coloring: \places \rightarrow \powerset(\types)$ is a place typing function;
%
\item $\inflow: \places \times \transitions \rightarrow \mult{\Omega_{\varset}}$ is an input flow, 
s.t. $\vartype(\inflow(p,t))=\coloring(p)$ for every $(p,t)\in\places\times\transitions$;
\item $\outflow: \transitions \times \places \rightarrow \mult{\tuples{\vars \cup \dom_\types}}$ is an output flow, s.t. $\vartype(\outflow(t,p))=\coloring(p)$ for every $(t,p)\in\transitions\times\places$;
%

\item $\guass:  \transitions \rightarrow \set{Q\land \varphi\mid Q\in\ucqd, \varphi\in\conds}$ is a partial guard assignment function, s.t., for every $\guass(t)=Q\land\varphi$ and $t\in\transitions$, the following holds:  
\begin{compactenum}
	\item $\varsin{\varphi}\subseteq\invars{t}$, where $\invars{t}=\cup_{p\in\places}\varsin{\inflow(p,t)}$;
	\item $(\varsin{Q}\cap \outvars{t})\setminus\invars{t}\neq\emptyset$ and $\varsin{Q}\subseteq \varsin{t}$, where $\outvars{t}=\cup_{p\in\places}\varsin{\outflow(t,p)}$ and $\varsin{t}=\invars{t}\cup\outvars{t}$.
\end{compactenum}

\end{compactenum}
\end{definition}

Here, the role of guards is twofold. On the one hand, similarly, for example, to CPNs, guards are used to impose \emph{conditions} (using $\varphi$) on tokens flowing through the net. On the other hand, a guard of a transition $t$ may also \emph{query} (using $Q$) the catalog in order to propagate some data into the net. The acquired data may be still filtered by using $\invars{t}$. Note that in condition \emph{(b)} of the guard definition we specify that there are some variables (excluding the fresh ones) in the outgoing arc inscriptions that do not appear in $\invars{t}$ and that are used by $Q$ to insert data from the catalog. Moreover, it is required that all free variables of $Q$ must coincide with the variables of inscriptions on outgoing and incoming arcs of a transition it is assigned to.
In what follows, we shall define arc inscriptions as $k\cdot\omega$, where $k\in\naturals$ and $\omega\in\Omega_A$ (for some set $A$).

\smallskip
\noindent
\textbf{Semantics.}
The execution semantics of a \clognet is similar to the one of CPNs. Thus, as a first step we introduce the standard notion of net marking. 
Formally, a \emph{marking} of a \clognet $N=(\types, \schema_\types, \places,\transitions,\inflow,\outflow,\coloring,\guass)$ is a function $m:\places\rightarrow \mult{\Omega_\types}$, so that  $m(p)\in\mult{\dom_{\coloring(p)}}$ for every $p\in\places$. 
We write $\markedcat{N}{m}{Cat}$ to denote \clognet $N$ marked with $m$, and equipped with a read-only catalog instance $Cat$ over $\schema_\types$. 

The firing of a transition $t$ in a marking is defined w.r.t.~a so-called \emph{binding} for $t$ defined as $\sigma : \varsin{t} \rightarrow \types$. Note that, when applied to (multisets of) tuples, $\sigma$ is applied to every variable singularly. 
For example, given $\sigma=\set{x\mapsto 1, y\mapsto \cname{a}}$, its application to a multiset of tuples $\omega=\set{\tup{x,y}^2,\tup{x,\cname{b}}}$ results in $\sigma(\omega)=\set{\tup{1,\cname{a}}^2,\tup{x,\cname{b}}}$.
\begin{definition}
\label{def:trans_e}
A transition $t \in\transitions$ is \emph{enabled} in a marking $m$ and a fixed catalog instance $Cat$, written $\enabledcat{m}{t}{Cat}$, if there exists binding $\sigma$ satisfying the following:
\begin{inparaenum}[\it (i)]
\item  $ \sigma(\inflow(p,t))\subseteq m(p)$, for every $p\in\places$;
\item $\sigma(\guass(t))$ is true;
\item $\sigma(x)\not\in \val(m)\cup \val(\cat)$, for
every $x\in\nuvarset \cap \outvars{t}$;\footnote{Here, with slight abuse of notation, we define by $\val(m)$ the set of all values appearing in $m$. }
\item $\sigma(x)\in\ans(Q,\cat)$ for $x\in\outvars{t}\setminus(\nuvarset\cup\invars{t})\cap\varsin{Q}$ and query $Q$ from $\guass(t)$ .
\end{inparaenum}
\end{definition}
Essentially, a transition is enabled with a binding $\sigma$ if the binding selects data objects carried by tokens from the input places and the read-only catalog instance, so that the data they carry make the guard attached to the transition true.

When a transition $t$ is enabled, it may fire. Next we define what are the effects of firing a transition with some binding $\sigma$.

\begin{definition}
\label{def:trans_f}
Let $\markedcat{N}{m}{Cat}$ be a marked \clognet, and $t\in\transitions$ a transition enabled in $m$ and $Cat$ with some binding $\sigma$. Then, $t$ may \emph{fire} producing a new marking $m'$, with $m'(p)=m(p)-\sigma(\inflow(p,t))+\sigma(\outflow(t,p))$ for every $p\in\places$. We denote this as $\firecat{m}{t}{m'}{Cat}$ and assume that the definition is inductively extended to sequences $\tau\in\transitions^*$.
\end{definition}
For $\markedcat{N}{m_0}{Cat}$ we use $\mathcal{M}(N)=\set{m\mid \exists \tau\in\transitions. \firecat{m_0}{\tau}{m}{Cat}}$ to denote the set of all markings of $N$ reachable from its initial marking $m_0$. 

Given $b\in\naturals$, a marked \clognet $\markedcat{N}{m_0}{Cat}$ is called \emph{bounded with bound $b$} if $|m(p)|\leq b$, for every marking $m\in\mathcal{M}(N)$ and every place $p\in\cplaces$. 

\smallskip
\noindent
\textbf{Execution semantics.}
The execution semantics of a marked \clognet $\markedcat{N}{m_0}{Cat}$ is defined in terms of a possibly infinite-state transition system in which states are labeled by reachable markings and each arc (or transition) corresponds to the firing of a transition in $N$ with a given binding. The transition system captures all possible executions supported by the net, by interpreting concurrency as interleaving. 
See Appendix~\ref{app:semantics} for the formal definition of how this transition system is induced.

As pointed out before, we are interested in analysing a \clognet irrespectively of the actual content of the catalog. Hence, in the following when we mention a (catalog-parameterised) marked net $\marked{N}{m_0}$ without specifying how the catalog is instantiated, we actually implicitly mean the \emph{infinite set of marked nets} $\markedcat{N}{m_0}{Cat}$ for every possible instance $Cat$ defined over the catalog schema of $N$.

We close with an example that illustrates all the main features of \clognets.

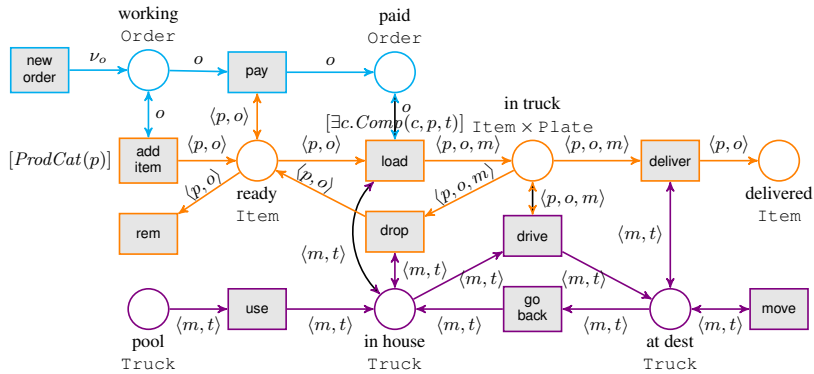
\begin{figure}[t!]

\tikzstyle{truck}=[draw=violet]
\tikzstyle{order}=[draw=cyan]
\tikzstyle{item}=[draw=orange]
\tikzstyle{mytrans}=[transition,font=\scriptsize,minimum width=10mm]

\newcommand{\order}{\datatype{Order}}
\newcommand{\barcode}{\datatype{BarCode}}
\newcommand{\oitem}{\datatype{Item}}
\newcommand{\plate}{\datatype{Plate}}
\newcommand{\truck}{\datatype{Truck}}

\centering
\resizebox{.9\textwidth}{!}{
\begin{tikzpicture}[->,>=stealth',auto,x=15mm,y=1.0cm,node distance=20mm and 15mm,thick]

\node[place,truck] (pool) {};
\node[below=-1mm of pool] {
  \begin{tabular}{@{}c@{}}
    pool\\
    $\truck$  
  \end{tabular}
};

\node[mytrans,right=10mm of pool,truck] (use) {\transition{use}};

\node[place,right=of use,truck] (inhouse) {};
\node[below=-1mm of inhouse] {
  \begin{tabular}{@{}c@{}}
    in house\\
    \truck  
  \end{tabular}
};

\node[mytrans,right=of inhouse,truck] (goback) {\begin{tabular}{@{}c@{}}\transition{go}\\\transition{back}\end{tabular}};

\node[mytrans,above=5mm of goback,truck] (drive) {\transition{drive}};

\node[place,right=of goback,truck] (atdest) {};
\node[below=-1mm of atdest] {
  \begin{tabular}{@{}c@{}}
    at dest\\
    \truck  
  \end{tabular}
};

\node[mytrans,right=10mm of atdest,truck] (move) {\transition{move}};

  \node[mytrans,above=18mm of pool,item] (ai) {
    \begin{tabular}{@{}c@{}}
      \transition{add}\\      
      \transition{item}
    \end{tabular}
  };
  \node[left=0mm of ai] {\footnotesize $\left[\prodcat(p)\right]$};
  
  \node[mytrans,below=5mm of ai,item] (remove) {\transition{rem}};
    
  \node[place,right=10mm of ai,item] (ready) {};
  \node[below=-1mm of ready] {
    \begin{tabular}{@{}c@{}}
      ready\\
      \oitem  
    \end{tabular}
  };
  
  \node[mytrans,right=of ready,item] (load) {\transition{load}};
  \node[above=0mm of load] (loadguard) {\footnotesize$[\exists c.\compatible(c,p,t)]$};
  
  \node[mytrans,below=5mm of load,item] (drop) {\transition{drop}};

  \node[place,right=of load,item] (intruck) {};
  \node[above=-1mm of intruck] {
    \begin{tabular}{@{}c@{}}
      in truck\\
      $\oitem\times\plate$  
    \end{tabular}
  };

  \node[mytrans,right=of intruck,item] (deliver) {\transition{deliver}};

  \node[place,right=10mm of deliver,item] (delivered) {};
  \node[below=-1mm of delivered] {
    \begin{tabular}{@{}c@{}}
      delivered\\
      \oitem  
    \end{tabular}
  };

  \node[place,above=8mm of ai,order] (inprep) {};
  \node[above=-1mm of inprep] {
    \begin{tabular}{@{}c@{}}
      working\\
      \order  
    \end{tabular}
  };

  \node[mytrans,left=10mm of inprep,order] (new) {
    \begin{tabular}{@{}c@{}}
    \transition{new}\\
    \transition{order}  
    \end{tabular}
  };

  \node[mytrans,above=8mm of ready,order] (pay) {\transition{pay}};  
  \node[place,right=of pay,order] (paid) {};
  \node[above=-1mm of paid] {
    \begin{tabular}{@{}c@{}}
      paid\\
      \order  
    \end{tabular}
  };

\draw[truck,->] (pool) --node[below]{$\tup{m,t}$} (use);
\draw[truck,->] (use) --node[below]{$\tup{m,t}$} (inhouse);
\draw[truck,->] (inhouse) --node[below,xshift=4mm,yshift=2mm]{$\tup{m,t}$} (drive);
\draw[truck,->] (drive) --node[below,xshift=-4mm,yshift=2mm]{$\tup{m,t}$} (atdest);
\draw[truck,->] (atdest) --node[below]{$\tup{m,t}$} (goback);
\draw[truck,->] (goback) --node[below]{$\tup{m,t}$} (inhouse);
\draw[truck,<->] (atdest) --node[below]{$\tup{m,t}$} (move);

\draw[->,item,sloped] (ready) --node[yshift=-1mm,above]{$\tup{p,o}$} (remove);
\draw[->,item] (ready) --node{$\tup{p,o}$} (load);
\draw[<->,truck,out=135,in=-135] (inhouse) edge node{$\tup{m,t}$} (load);

\draw[->,item,sloped] (intruck) --node[above,yshift=-1mm]{$\tup{p,o,m}$} (drop);
\draw[->,item,sloped] (drop) --node[above,yshift=-1mm,xshift=-2mm]{$\tup{p,o}$} (ready);
\draw[<->,truck] (inhouse) --node[right]{$\tup{m,t}$} (drop);

\draw[->,item] (load) --node{$\tup{p,o,m}$} (intruck);
\draw[->,item] (intruck) --node{$\tup{p,o,m}$} (deliver);
\draw[<->,item] (drive) edge node[right]{$\tup{p,o,m}$} (intruck);

\draw[<->,truck] (atdest) --node{$\tup{m,t}$} (deliver);
\draw[->,item] (deliver) --node{$\tup{p,o}$} (delivered);

\draw[->,order] (new) --node{$\nu_o$} (inprep);
\draw[->,order] (inprep) --node{$o$} (pay);
\draw[->,order] (pay) --node{$o$} (paid);

\draw[<->,order] (inprep) --node{$o$} (ai);
\draw[->,item] (ai) --node{$\tup{p,o}$} (ready);
\draw[<->,item] (ready) --node{$\tup{p,o}$} (pay);
\draw[<->,order,out=90,in=-90] (load) edge node[right,pos=.7]{$o$} (paid);
  
\end{tikzpicture}
}
\caption{A \clognet (its catalog is in Example~\ref{ex:catalog}). In the picture, $\oitem$ and $\truck$ are compact representations for $\producttype \times \order$ and $\plate \times \trucktype$ respectively. The top blue part refers to orders, the central orange part to items, and the bottom violet part to delivery trucks.}
\label{fig:clognet}
\end{figure}

\begin{example}
\label{ex:clognet}
Starting from the catalog in Example~\ref{ex:catalog}, Figure~\ref{fig:clognet} shows a simple, yet sophisticated example of \clognet capturing the following order-to-delivery process. Orders can be created by executing the \transition{new order} transition, which uses a $\nu$-variable to generate a fresh order identifier. A so-created, working order can be populated with items, whose type is selected from those available in the catalog relation $\prodcat$. Each item then carries its product type and owning order. When an order contains at least one item, it can be paid. Items added to an order can be removed or loaded in a compatible truck. The set of available trucks, indicating their plate numbers and types, is contained in a dedicated \emph{pool} place. Trucks can be borrowed from the pool and placed in house. An item can be loaded into a truck if its owning order has been paid, the truck is in house, and the truck type and product type of the item are compatible according to the $\compatible$ relation in the catalog. Items (possibly from different orders) can be loaded in a truck, and while the truck is in house, they can be dropped, which makes them ready to be loaded again. A truck can be driven for delivery if it contains at least one loaded item. Once the truck is at its destination, some items may be delivered (this is simply modelled non-deterministically). The truck can then either move, or go back in house. 
\end{example}

Example~\ref{ex:clognet} shows various key aspects related to modelling data-aware processes with multiple case objects using \clognets. First of all, whenever an object is involved in a many-to-one relation from the ``many'' side, it then becomes responsible of carrying the object to which it is related. This can be clearly seen in the example, where each item carries a reference to its owning order and, once loaded into a truck, a reference to the truck plate number. 
Secondly, the three object types involved in the example show three different modelling patterns for their creation. Unboundedly many orders can be genuinely created using a $\nu$-variable to generate their (fresh) identifiers. The (finite) set of trucks available in the domain is instead fixed in the initial marking, by populating the \emph{pool} place. The \clognet shows that such trucks are used as resources that can change state but are never destroyed nor created. Finally, the case of items is particularly interesting. Items in fact can be arbitrarily created and destroyed. However, their creation is not modelled using an explicit $\nu$-variable, but is instead simply obtained by the \transition{add item} transition with the usual token-creation mechanism of standard (non-coloured) Petri nets. Thanks to the multiset semantics of Petri nets, it is still possible to create multiple items having the same product type and owning order. However, it is not possible to track the evolution of a specific item, since there is no explicit identifier carried by item tokens. This is not a limitation in this example, since items are not referenced by other objects present in the net (which is instead the case for orders and trucks). All in all, this shows that $\nu$-variables are only necessary when the \clognet needs to handle the arbitrary creation of objects that are referenced by other objects.

%

%% file: 4-translation.tex
\newcommand{\mcinline}[1]{\texttt{#1}}
\section{From \clognets to MCMT}
\label{sec:translation}
We now report on the encoding of \clognets into the verification language supported by the \mcmt model checker, showing that the various modelling constructs of \clognets have a direct counterpart in \mcmt, and in turn enabling formal analysis. 

\mcmt is founded on the theory of \emph{array-based systems}, an umbrella term used to refer to \emph{infinite-state transition systems} specified using a declarative,
logic-based formalism by which arrays are manipulated 
via logical updates. 
An array-based system is represented using a multi-sorted
theory with two sorts: one for the indexes of
arrays, and the other for the elements stored therein. Since the content of an
array changes over time, it is referred to by a function
variable, whose interpretation in a state is that of a total function mapping
indexes to elements (applying the function to an index denotes the
classical \emph{read} array operation).
We adopt here the module of \mcmt called ``database-driven applications'', which supports the representation of read-only databases.
%
%
%

Specifically, we show how to encode a \clognet $\marked{N}{m_0}$, where $N=(\types, \schema_\types, \places,\transitions,\inflow,\outflow,\coloring,\guass)$ into (data-driven) \mcmt specification. The translation is split into two phases. First, we tackle the type domain and catalog. Then, we present a step-wise encoding of the \clognet places and transitions into arrays.

\smallskip
\noindent
\textbf{Data and schema translation.} We start by describing how to translate static data-related components.  Let  $\types=\set{\type_1,\ldots,\type_{n_d}}$. Each data type $\type_i$ is encoded in \mcmt with declaration \mcinline{:smt (define\_type Di)}.
For each declared type $\type$ \mcmt implicitly generates a special \texttt{NULL} constant indicating an empty/undefined value of $\type$.

 To represent the catalog relations of $\schema_\types=\set{R_1,\ldots,R_{n_r}}$ in \mcmt, we proceed as follows. Recall that in catalog every relation schema has $n+1$ typed attributes among which some may be foreign keys referencing other relations, its first attribute is a primary key, and, finally, primary keys of different relation schemas have different types. With these conditions at hand, we adopt the functional characterisation of read-only databases studied in \cite{MSCS20}. For every relation $R_i(id,A_1,\ldots,A_n)$ with $\PK{R}=\set{id}$, we introduce unary functions that correctly reference each attribute of $R_i$ using its primary key. More specifically, for every $A_j$ ($j=1,\ldots,n$) we create a function $f_{R_i,A_j}:\dom_{\vartype(id)}\rightarrow\dom_{\vartype{A_j}}$. If $A_j$ is referencing an identifier of some other relation $S$ (i.e., $\FK{R_i}{A_j}{S}{id}$), then $f_{R_i,A_j}$ represents the foreign key referencing to $S$. Note that in this case the types of $A_j$ and ${S}.{id}$ should coincide. In \mcmt, assuming that $\texttt{id\_D}=\vartype(id)$ and $\texttt{Aj\_D}=\vartype(A_j)$, this is captured using statement \mcinline{:smt (define Ri\_Aj ::(-> id\_D Aj\_D))}.

All the constants appearing in the net specification must be properly defined. Let $C=\set{v_1,\ldots,v_{n_c}}$ be the set of all constants appearing in $N$. $C$ is defined as $\bigcup_{t\in\transitions}\constin{\guass(t)}\cup\supp{m_0}\cup\bigcup_{t\in\transitions,p\in\places} \constin{\outflow(t,p)}$. Then, every constant $v_i\in C$ of type $\type$ is declared in \mcmt as \mcinline{:smt (define vi ::D)}.



To make \mcmt aware of the fact that these elements have been declared to describe a read-only database schema, the following code section is needed:
\begin{wrapfigure}[4]{l}{60mm}
\vspace*{-1cm}
\begin{mc}
:db_driven
:db_sorts D1,...,Dnd
:db_functions R1_A1,...,Rnr_Ak
:db_constants v1,...,vnc
\end{mc}
\end{wrapfigure}

\noindent
\textbf{Places.} Given that, during the net execution, every place may store unboundedly many tokens, we need to ensure a potentially infinite provision of values to $p$ using unbounded arrays. 
To this end, every place $p\in\places$ with $\coloring(p)=\type_1\times\ldots\times\type_k$ is going to be represented as a combination of arrays $p_1,\ldots,p_k$, where a special index type $\typename{P_{ind}}$ (disjoint from all other types) with domain $\dom_{P_{ind}}$  is used as the array index sort and $\type_1,\ldots,\type_k$ account for the respective target sorts of the  arrays.\footnote{in \mcmt  there is no need to declare them explicitly.} In \mcmt, this is declared as
\mcinline{:local p\_1 D1    ...   :local p\_k Dk}.
Then, intuitively, we associate to the $j$-th token $(v_1,\ldots,v_k)\in m(p)$ an element $j\in\dom_{P_{ind}}$ and a tuple $(j,p_1[j],\ldots,p_k[j])$, where $p_1[j]=v_1,\ldots, p_k[j]=v_k$. Here, $j$ is an \emph{``implicit identifier''} of this tuple in $m(p)$. Using this intuition and assuming that there are in total $n$ control places, we  represent the initial marking $m_0$ in two steps (a direct declaration is not possible due to the language restrictions of \mcmt). First, we symbolically declare that all places are by default empty
using the \mcmt initialisation statement:
\begin{wrapfigure}[5]{l}{20mm}
\vspace*{-1cm}
\begin{mc}
:initial   
:var x
:cnj $init\_p_1$ 
     ... 
     $init\_p_n$
\end{mc}
\end{wrapfigure}

\noindent Here, \texttt{cnj} represents a conjunction of atomic equations that, for ease of reading, we organized in blocks, where each $init\_p_i$ specifies for place $p_i\in\places$ with $\coloring(p_i)=\type_1\times\ldots\times \type_k$ that it contains no tokens. This is done by explicitly ``nullifying'' all component of each possible token in $p_i$, written in \mcmt  as 
\mcinline{(= pi\_1[x] NULL\_D1)(= pi\_2[x] NULL\_D2)...(= pi\_k NULL\_DK)}.
The initial marking is then injected with a dedicated \mcmt code that populates the empty place arrays with records representing the tokens therein. Due to lack of space, the  code is shown in Appendix~\ref{app:initial-marking-mcmt}.




\begin{wrapfigure}[7]{l}{4.8cm}
\centering
\vspace*{-8mm}
\resizebox{4.85cm}{!}{
\begin{tikzpicture}[->,>=stealth',auto,x=1cm,y=1.0cm,thick]

   \node[transition] (t) at (2,0) {~~~~\large $t$~~~~};
   	\node[above of = t, yshift=-.8cm,label=\large $\mathbf{[}{\color{mgreen}Q\land \varphi}\mathbf{]}$]  (guard)  {};

	\node[place,left of=t,xshift=-2cm,yshift=1.2cm,label={[yshift=-0.07cm]above:\large $pin_1$}] (pin_1) {};
	\node[place,left of=t,xshift=-2cm,yshift=-0cm,label={[yshift=0.07cm]below:\large $pin_k$}] (pin_k) {};
	\node[left of=t,xshift=-2cm,yshift=.3cm,label=\tiny{$\bullet \bullet \bullet$}]  (dots2) {};

	\node[place,right of=t,xshift=2cm,yshift=1.2cm,label={[yshift=-0.07cm]above:\large $pout_1$}] (pout_1) {};
	\node[place,right of=t,xshift=2cm,yshift=-0cm,label={[yshift=0.07cm,xshift=0.07cm]below:\large $pout_n$}] (pout_n) {};
	\node[right of=t,xshift=2cm,yshift=.3cm,label=\tiny{$\bullet \bullet \bullet$}]  (dots3) {};

  \path[]
    (pin_1) edge node[above,xshift=-1mm,sloped] {\large $\cname{ri_1}\cdot\vec{in_1}$} (t)    
    (pin_k) edge node[below,xshift=-1mm,sloped] {\large $\cname{ri_k}\cdot\vec{in_k}$} (t)
    (t) edge node[above,xshift=-1mm,sloped, pos=.6] {\large $\cname{ro_1}\cdot\vec{out_1}$} (pout_1)    
    (t) edge node[below,xshift=-1mm,sloped,pos=.55] {\large $\cname{ro_n}\cdot\vec{out_n}$} (pout_n)
    ;
 
\end{tikzpicture}}
\caption{A generic \clognet transition ($\cname{ri_j}$ and $\cname{ro_j}$ are  natural numbers)}
\label{fig:t-generic}
\end{wrapfigure}

\smallskip
\noindent
\textbf{Transition enablement and firing.} 
We now show how to check for transition enablement and compute the effect of a transition firing in \mcmt. To this end, we consider the generic, prototypical \clognet transition $t\in\transitions$ depicted in Figure~\ref{fig:t-generic}. The enablement of this transition is subject to the following conditions: 
\begin{inparaenum}
\item[\textit{(FC1)}] there is a binding $\sigma$ that correctly matches tokens in the places to the corresponding inscriptions on the input arcs (i.e., each place $pin_i$ provides enough tokens required by a corresponding inscription $F(pin_i,t)=\vec{in_i}$), and that computes new and possibly \emph{fresh} values that are pairwise distinct from each other as well as from all other values in the marking;
\item[\textit{(FC2)}] the guard $\guass(t)$ is satisfied under the selected binding.
\end{inparaenum} 
In \mcmt, $t$ is captured with a transition statement consisting of a guard $G$ and an update $U$:
\begin{wrapfigure}[5]{l}{50mm}
\vspace*{-10mm}
\begin{mc}
:transition
:var x,x1,...,xK,y1,...,yN
:var j
:guard $G$
... $U$ ...
\end{mc}
\end{wrapfigure}

\noindent Here every \texttt{x} (resp., \texttt{y}) represents an existentially quantified index variables corresponding to variables in the incoming inscriptions (resp., outgoing inscriptions), $\texttt{K}=\sum_{j\in\set{1,\ldots,k}}\texttt{ri}_j$, $\texttt{N}=\sum_{j\in\set{1,\ldots,n}}\texttt{ro}_j$ and \texttt{j} is a universally quantified variable, that will be used for computing bindings of $\nu$-variables.
In the following we are going to elaborate on the construction of the \mcmt transition statement. We start by discussing the structure of $G$ which in \mcmt is represented as a conjunction of atoms or negated atoms and, intuitively, should address all the conditions stated above. 


First, to construct a binding that meets condition \textit{(FC1)}, we need to make sure that every place contains enough of tokens that match a corresponding arc inscription. Using the array-based representation, for every place $pin_i$ with $\inflow(pin_i,t)=\cname{ri_1}\cdot\vec{in_i}$ and $|\coloring(pin_i)|=k$, we can check this with a formula 
\[
\psi_{pin_i}:=
  \exists \texttt{x}_1,\ldots,\texttt{x}_{\cname{ri_i}}.
  \hspace*{-1cm}\bigwedge_{
    \substack{j_1,j_2\in\set{\texttt{x}_1,\ldots,\texttt{x}_\cname{ri_j}},j_1\neq j_2,\\l\in\set{1,\ldots,k}}} 
  \hspace*{-1cm}
  pin_{i,l}[j_1]=pin_{i,l}[j_2]\land \bigwedge_{l\in\set{1,\ldots,k}} pin_{i,l}[\texttt{x}_1]\neq \texttt{NULL\_D}_l
\]
Given that variables representing existentially quantified index variables are already defined, in \mcmt this is encoded as conjunctions of atoms \texttt{(= pini\_$l$[$j_1$] pini\_$l$[$j_2$])} and atoms \texttt{not(= pini\_$l$[x1] NULL\_D$l$)}, where \texttt{NULL\_D$l$} is a special null constant of type of elements stored in \texttt{pini\_$l$}. All such conjunctions, for all input places of $t$, should be appended to $G$. 

We now define the condition that selects proper indexes in the output places so as to fill them with the tokens generated upon transition firing. To this end, we need to make sure that arrays corresponding to the output places of $t$ (as well as all the $m$ arrays of the system) contain no values in the slots marked by \texttt{y} index variables. This is represented using  a formula
\begin{equation}\label{eq:out}
\psi_{pout_i}:=
  \exists \texttt{y}_1,\ldots,\texttt{y}_{\cname{ri_i}}.
  \bigwedge_{{j\in\set{\texttt{y}_1,\ldots,\texttt{y}_\cname{ri_j}},\\ s\in\set{1,\ldots,m},l\in\set{1,\ldots,k}}}  pout_{s,l}[j]=\texttt{NULL\_D}_l,
\end{equation}
which is encoded in \mcmt similarly to the case of $\psi_{pin_i}$.
The fact that it is necessary to check the \texttt{y}-index for \emph{all} the arrays of the system is due to the following MCMT technicality.  Indeed, in the setting of \cite{MSCS20}, many ``index'' sorts are allowed, whereas in MCMT specifications we have only one such sort at our
disposal. There is no loss of generality in that, because all the transitions can be specified so as to maintain a
suitable invariant. For instance, if there are two places $p1$ and $p2$, with, say, array components $p1_1, p1_2$
and $p2_1, p2_2, p2_3$ respectively, then the system must be designed so as to maintain the invariant
$$\forall j \bigwedge_{l,m} (p1_l[j]=NULL \lor p2_m[j]=NULL)$$
This invariant allows to associate with every non-null entry a unique index sort, i.e. a unique place. The format of Formula \eqref{eq:out} guarantees that the invariant is maintained. 

Moreover, when constructing a binding, we have to take into account the case of arc inscriptions causing implicit ``joins'' between the net marking and data retrieved from the catalog. This happens when there are some variables in the input flow that coincide with variables of $Q$, i.e., $\varsin{\inflow(pin_j,t)}\cap\varsin{Q}\neq\emptyset$. For ease of presentation, denote the set of such variables as $\mathbf{s}=\set{s_1,\ldots, s_r}$ and introduce a function $\pi$ that returns the position of a variable in a tuple or relation. E.g., $\pi(\tup{x,y,z},y)=2$, and $\pi(R,B)=3$ in $R(id,A,B,E)$.
Then, for every relation $R$ in $Q$ we generate a formula 
\[
\psi_R:=\bigwedge_{{j\in\set{1,\ldots,k}, s\in\bigl(\mathbf{s}\cap\varsin{R}\bigr)}}pin_{j,{\pi(\vec{in_j},s)}}[\texttt{x}]=f_{R,A_{\pi(R,s)}}(id)
\]
This formula guarantees that values provided by a constructed binding respect the aforementioned case for some index \texttt{x} (that has to coincide with one of the index variables from $\psi_{pin_j}$) and identifier $id$.  In \mcmt this is encoded as a conjunction of atoms \texttt{(= (R\_Ai id) pinj\_l[x])}, where $\texttt{i}=\pi(R,s)$ and $\texttt{l}=\pi(\vec{in_j},s)$. As in the previous case, all such formulas are appended to $G$.

We now incorporate the encoding of condition \textit{(FC2)}.
Every variable $z$ of $Q$ with $\vartype(z)=\texttt{D}$ has to be declared in \mcmt as \texttt{:eevar z D}.
We call an \emph{extended guard} a guard $Q^e\land\varphi^e$ in which every relation $R$ has been substituted with its functional counterpart and every variable $z$ in $\varphi$ has been substituted with a ``reference'' to a corresponding array $pin_j$ that $z$ uses as a value provider for its bindings. More specifically, every relation $R/n+1$ that appears in $Q$ as $R(id,z_1,\ldots,z_n)$  is be replaced by  conjunction $id\neq \texttt{NULL\_D}\land f_{R,A_1}(id)=z_1 \land \ldots \land f_{R,A_n}(id)=z_n$, where $\texttt{D}=\vartype(id)$. In \mcmt, this is written as \mcinline{(not (= id NULL\_D)) $expr_1$ ... $expr_n$}.
Here, every $expr_i$ corresponds to an atomic equality from above and is specified in \mcmt in three different ways based on the nature of $z_i$. Let us assume that $z_i$ has been declared before as \texttt{:eevar z\_1 D}. If $z_i$ appears in a corresponding incoming transition inscription, then $expr_i$ is defined as \texttt{(= (R\_Ai id) pin\_j[x])(= z\_i pin\_j[x])}, where \texttt{i}-th attribute of $R$ coincides with the \texttt{j}-th variable in the inscription $\inflow(pin,t)$.
If $z_i$ is a variable bound by an existential quantifier in $Q$, then $expr_i$ in \mcmt is going to look as \texttt{(= (R\_Ai id) zi)}. Finally, if $z_i$ is a variable in an outgoing inscription used for propagating data from the catalog  (as discussed in condition \emph{(1)}), then $expr_i$ is simply defined with the following statement: \texttt{(not(= z\_i NULL\_Di))(= (R\_Ai id) z\_i)}, where \texttt{Di} is the type of $z_i$. 

We now consider the assignment of a propagated data.
Variables in $\varphi$ are substituted with their array counterparts. In particular, every variable $z\in\varsin{\varphi}$ is substituted with \texttt{pinj\_i[x]}, where $\texttt{i}=\pi(\vec{in_j},z)$. Given that $\varphi$ is represented as a conjunction of variables, its representation in \mcmt together with the aforementioned substitution is trivial. To finish the construction of $G$, we append to it the \mcmt version of $Q^e\land\varphi^e$.

We come back to condition \textit{(FC1)} and show how bindings are generated for $\nu$ variables of the output flow of $t$. 
In \mcmt we use a special universal guard \texttt{:uguard} (to be inserted right after the \texttt{:guard} entry) that, for every variable $\nu\in\nuvarset\cap(\outvars{t}\setminus\varsin{\vec{out_j}})$ previously declared using \texttt{:eevar nu D}, and for arrays $p_1,\ldots,p_k$ with target sort \texttt{D}, consists of expression
\mcinline{(not(=nu p\_1[j]))...(not(=nu p\_k[j]))}.
This encodes ``local'' freshness for $\nu$-variables, which suffice for our goal.

After a binding has been generated and the guard of $t$ has been checked, a new marking is generated by assigning corresponding tokens to the outgoing places and by removing tokens from the incoming ones. Note that, while the tokens are populated by assigning their values to respective arrays, the token deletion happens by nullifying (i.e., assigning special \texttt{NULL} constants) entries in the arrays of the input places. All these operations are specified in the special update part of the transition statement $U$ and are captured in \mcmt as:
\begin{wrapfigure}[7]{l}{25mm}
\vspace*{-10mm}
\begin{mc}
:numcases NC
...
:case (= j $i$)
:val $v_{1,i}$
...
:val $v_{k,i}$
...
\end{mc}
\end{wrapfigure}
Here, the transition runs through \texttt{NC} cases. All the following cases go over the indexes \texttt{y1},\ldots, \texttt{yN} that correspond to tokens that have to be added to places.
More specifically, for every place $pout\in\places$ such that $|\coloring(pout)|=k$, we add an $i$-th token to it by putting a value $v_{r,i}$ in $i$-th place of every $r$-th component array of $pout$. This $v_{r,i}$ can either be a $\nu$-variable \texttt{nu} from the universal guard, or a value coming from a place $pin$ specified as \texttt{pin[xm]} (from some \texttt{x} input index variable) or a value from some of the relations specified as \texttt{(R\_Ai id)}. Note that \texttt{id} should be also declared as \texttt{:eevar id id\_D}, where $\vartype(\texttt{id})=\texttt{id\_D}$.
Every \texttt{:val v} statement follows the order in which all the local and global variables have been defined, and, for array variables  $a$ and every every case \texttt{(= j $i$)}, such statement stands for a simple assignment $a[i]:=\texttt{v}$.

%
%

\smallskip\noindent
\textbf{Implementation status.} The provided translation is fully compliant with the concrete specification language \mcmt. The current implementation has however a limitation on the number of supported index variables in each \mcmt transition statement. Specifically, two existentially quantified and one universally quantified variables are currently supported. This has to be taken into account if one wants to run the model checker on the result produced by translating a \clognet, and possibly requires to rewrite the net (if possible) into one that does not exceed the supported number of index variables.

This limitation is not dictated by algorithmic nor theoretical limitations, but is a mere characteristic of the current implementation, and comes from the fact that the wide range of systems verified so far with \mcmt never required to simultaneously quantify on many array indexes. 
%
%
There is an ongoing implementation effort for a new version of \mcmt that supports arbitrarily many quantified index variables, and consequently concrete model checking of the full \clognet model is at reach.


%% file: verification.tex
\newcommand{\bsearch}{\textsc{\footnotesize BReach}}
\newcommand{\crorder}{\relname{CrOrder}}
\section{Parameterised Verification}

Thanks to the encoding of \clognets into (the data-driven module of) \mcmt, we can handle the parameterised verification of safety properties over \clognets, and study crucial properties such as soundness, completeness, and termination by relating \clognets with the foundational framework underlying such an \mcmt module \cite{MSCS20,lmcs}.

 This amounts to verifying whether it is true that \emph{all} the reachable states of a marked \clognet satisfy a desired condition, \emph{independently from the content of the catalog}. As customary in this setting, this form of verification is tackled in a converse way, by formulating an \emph{unsafe condition}, and by checking whether there exists an instance of the catalog such that the \clognet can evolve the initial marking to a state where the unsafe condition holds. Technically, given a property $\psi$ capturing an unsafe condition and a marked \clognet $\marked{N}{m_0}$, we say that $\marked{N}{m_0}$ is \emph{unsafe} w.r.t.~$\psi$ if there exists a catalog instance $Cat$ for $N$ such that the marked \clognet with fixed catalog  $\markedcat{N}{m_0}{Cat}$ can reach a configuration where $\psi$ holds.

With a slight abuse of notation, we interchangeably use the term \clognet to denote the input net or its \mcmt encoding. We start by defining (unsafety) properties, in a way that again guarantees a direct encoding into the \mcmt model checker.

\begin{definition}
A \emph{property} over \clognet $N$ 
is a formula of the form $\exists \vec{x}.\psi$, where $\psi$ is a quantifier-free query that additionally contains atomic predicates $[p\geq c]$ and $[p(x_1,\ldots,x_n)\geq c]$, where $p$ is a place name of $N$, $c\in\naturals$, and $\varsin{\psi}=X_P$, with $X_P$ the set of variables appearing in all the atomic predicates $[p(x_1,\ldots,x_n)\geq c]$.
\end{definition}
Here, $[p\geq c]$ specifies that in place $p$ here are at least $c$  tokens. Similarly, $[p(x_1,\ldots,x_n)\geq c]$ indicates that in place $p$ there are at least $c$ tokens carrying the tuple $\tup{x_1,\ldots, x_n}$ of data objects. A property may also mention relations from the catalog, provided that all variables used therein also appear in atoms that inspect places. 

This can be seen as a language to express \emph{data-aware coverability properties} of a \clognet, possibly relating tokens with the content of the catalog. Focusing on covered markings as opposed as fully-specified reachable markings is customary in data-aware Petri nets or, more in general, well-structured transition systems (such as $\nu$-PNs \cite{RVFE11}).

\begin{example}
  Consider the \clognet of Example~\ref{ex:clognet}, with an initial marking that populates the $\mathit{pool}$ place with available trucks. Property $\exists p,o.[\mathit{delivered}(p,o) \geq 1] \land [\mathit{working}(o) \geq 1]$ captures the undesired situation where a delivery occurs for an item that belongs to a working (i.e., not yet paid) order. This can never happen, irrespectively of the content of the net catalog: items can be delivered only if they have been loaded in a compatible truck, which is possible only if the order of the loaded item is $\mathit{paid}$.
\end{example}

In the remainder of the section, we focus on the key properties of soundness, completeness and termination of the backward reachability procedure encoded in \mcmt, which can be used to handle the parameterised verification problem for \clognets defined above.\footnote{\emph{Backward reachability} is not \emph{marking reachability}. We consider reachability of a configuration satisfying a property that captures the covering of a data-aware marking.} We call this procedure $\bsearch$, and in our context we assume it takes as input a marked \clognet and an (undesired) property $\psi$, returning \unsafe if there exists an instance of the catalog so that the net can evolve from the initial marking to a configuration that satisfies $\psi$, and \safe otherwise. For details on the procedure itself, refer to~\cite{MSCS20,lmcs}. We characterise the (meta-)properties of this procedure as follows.

%
%
%

\begin{definition}
Given a marked \clognet $\tup{N,m_0}$ and a property $\psi$, $\bsearch$  is: 
\begin{inparaenum}[\it (i)]
\item \emph{sound} if, whenever it terminates, it produces a correct answer;
\item \emph{partially sound} if a \safe result it returns is always correct;
\item \emph{complete} (w.r.t.~unsafety) if, whenever $\tup{N,m_0}$ is \unsafe with respect to $\psi$, then $\bsearch$ detects it and returns \unsafe.
\end{inparaenum}
\end{definition}
In general, $\bsearch$ is not guaranteed to terminate (which is not surprising given the expressiveness of the framework and the type of parameterised verification tackled). 

As we have seen in Section~\ref{sec:translation}, the encoding of fresh variables requires to employ a limited form of universal quantification. This feature goes beyond the foundational framework for (data-driven) \mcmt \cite{MSCS20}, which in fact does not explicitly consider fresh data injection. It is known from previous works (see, e.g., \cite{AGPR12}) that when universal quantification over the indexes of an array is employed, $\bsearch$ cannot guarantee that all the indexes are considered, leading to potentially spurious situations in which some indexes are  simply ``disregarded'' when exploring the state space. This may wrongly classify a \safe case as being \unsafe, due to spurious exploration of the state space, similarly to what happens in lossy systems. By combining \cite{MSCS20} and \cite{AGPR12}, we then obtain: 

\begin{theorem}
$\bsearch$ is partially sound and complete for marked \clognets.
\end{theorem}

Fortunately, \mcmt is equipped with techniques \cite{AGPR12} for debugging the returned result, and tame partial soundness. In fact, \mcmt warns when the produced result is provably correct, or \emph{may} have been produced due to a spurious state-space exploration.  

A key point is then how to tame partial soundness towards recovering full soundness and completeness (and, possibly, termination). We obtain this by either assuming that the \clognet of interest does not employ at all fresh variables, or is bounded.

\smallskip\noindent
\textbf{Conservative \clognets} are \clognets that do not employ $\nu$-variables in arc inscriptions. It turns out that such nets are fully compatible with the foundational framework in \cite{MSCS20}, and consequently inherit all the properties established there. In particular, we obtain that $\bsearch$ is a semi-decision procedure.
\begin{theorem}
\label{thm:conservative}
$\bsearch$ is sound and complete for marked, conservative \clognets.
\end{theorem}

One may wonder whether studying conservative nets is meaningful. We argue in favour of this by considering modelling techniques to ``remove'' fresh variables present in the net. 
The first technique is to ensure that $\nu$-variables are used only when necessary. As we have extensively discussed at the end of Section~\ref{sec:clognets}, this is the case only for objects that are referenced by other objects. This happens when an object type participates on the ``one'' side of a many-to-one relationship, or for one of the two end points of a one-to-one relationship. The second technique is to limit the scope of verification by singling out only one (or a bunch of) ``prototypical'' object(s) of a given type. This is, e.g., what happens when checking soundness of workflow nets, where only the evolution of a single case from the input to the output place is studied. 

\begin{example}
We can turn the \clognet of Example~\ref{ex:clognet} into a conservative one by removing the \transition{new order} transition, and by ensuring that in the initial marking one or more order tokens are inserted into the $\mathit{working}$ place. This allows one to verify how these orders co-evolve in the net. A detected issue carries over the general setting where orders can be arbitrarily created.
\end{example}

A third technique is to remove the part of the \clognet with the fresh objects  creation, assuming instead that such objects are all ``pre-created'' and then listed in a read-only, catalog relation. This is more powerful than the first technique from above: now verification considers all possible configurations of such objects as described by the catalog schema. In fact, using this technique on Example~\ref{ex:clognet} we can turn the \clognet into a conservative \clognet that mimics exactly the behaviour of the original one.

\begin{example}
We make the \clognet from Example~\ref{ex:clognet} conservative in a way that reconstructs the original, arbitrary order creation. To do so we extend the catalog with a unary relation schema $\crorder$ accounting for (pre-)created orders. Then, we modify the \transition{new order} transition: we substitute the $\nu$-variable $\nu_o$ with a normal variable $o$, and we link this variable to the catalog, by adding as a guard a query $\crorder(o)$. This modified \transition{new order} transition extracts an order from the catalog and making it \emph{working}. Since in the original \clognet the creation of orders is unconstrained, it is irrelevant for verification if all the orders involved in an execution are created on-the-fly, or all created at the very beginning. Paired with the fact that the modified \clognet is analysed for all possible catalog instances, i.e., all possible sets of pre-created orders, this tells us that the original and modified nets capture the same relevant behaviours.
\end{example}

\smallskip\noindent
\textbf{Bounded \clognets}. An orthogonal approach is to study what happens if the \clognet of interest is bounded (for a given bound). In this case, we can ``compile away'' fresh-object creation by introducing a place that contains, in the initial marking, enough provision of pre-defined objects. This effectively transforms the \clognet into a conservative one, and so Theorem~\ref{thm:conservative} applies. 
If we consider a boudned \clognet  and its catalog is acyclic (i.e., its foreign keys cannot form referential cycles where a table directly or indirectly refers to itself), then it is possible to show using the results from~\cite{MSCS20} that verifying safety of conservative \clognets becomes decidable.

%
%

Several modelling strategies can be adopted to turn an unbounded \clognet into a bounded one. We illustrate two strategies in the context of our running example.

\begin{example}
Consider again the \clognet of Example~\ref{ex:clognet}. It has two sources of unboundedness: the creation of orders, and the addition of items to working orders. The first can be tackled by introducing suitable resource places. E.g., we can impose that each order is contolled by a manager and can be created only when there is an idle manager not working on any other order. This makes the overall amount of orders unbounded over time, but bounded in each marking by the number of resources. Items creation can be bounded by imposing, conceptually, that each order cannot contain more than a maximum number of items. This amounts to impose a maximum multiplicity on the ``many'' side of each one-to-many relation implicitly present in the \clognet.
\end{example}

%% file: 3-comparison.tex
\section{Comparison to Other Models}
   
We comment on how the formalism of \clognets relates to the most recent data-aware Petri net-based models, arguing that it provides an interesting mix of their main features. Due to space reasons, we discuss the comparison conceptually.


\smallskip
\noindent
\textbf{\dbnets.} \clognets in their full generality match with an expressive fragment of the \dbnet model \cite{MonR17}. \dbnets combine a control-flow component based on CPNs with fresh value injection a l\`a $\nu$-PNs with an underlying read-write persistent storage consisting of a relational database with full-fledged constraints. Special ``view'' places in the net are used to inspect the content of the underlying database, while transitions are equipped with database update operations. 

In \clognets, the catalog accounts for a persistent storage solely used in a ``read-only'' modality, thus making the concept of view places rather unnecessary. More specifically, given that the persistent storage can never be changed but only queried for extracting data relevant for running cases, the queries from view places in \dbnets have been relocated to transition guards of \clognets. While \clognets do not come with an explicit, updatable persistent storage, they can still employ places and suitably defined subnets to capture read-write relations and their manipulation. In particular, as shown in \cite{MonR19}, read-write relations queried using $\ucqd$ queries can be directly encoded with special places and transitions at the net level. The same applies to \clognets.
 
While verification of \dbnets has only been studied in the bounded case, \clognets are formally analysed here without imposing boundedness, and parametrically w.r.t.~read-only relations. In addition, the \mcmt encoding provided here constitutes the first attempt to make this type of nets practically verifiable.

\smallskip
\noindent
\textbf{PNIDs.} The net component of our \clognets model is equivalent to the formalism of Petri nets with identifiers (PNIDs \cite{HSVW09}) without inhibitor arcs. Interestingly, PNIDs without inhibitor arcs form the formal basis of the \emph{Information Systems Modelling Language} (ISML) defined in~\cite{PWOB19}. In ISML, PNIDs are paired with special CRUD operations to define how relevant facts are manipulated. Such relevant facts are structured according to a conceptual data model specified in ORM, which imposes structural, first-order constraints over such facts. This sophistication only permits to formally analyse the resulting formalism by bounding the PNID markings and the number of objects and facts relating them.  The main focus of ISML is in fact more on modelling and enactment. \clognets can be hence seen as a natural ``verification'' counterpart of ISML, where the data component is structured relationally and does not come with the sophisticated constraints of ORM, but where parameterised verification is practically possible.


\smallskip
\noindent
\textbf{Proclets.} \clognets can be seen as a sort of \emph{explicit data} version of (a relevant fragment of) Proclets \cite{Fahland19}. Proclets handle multiple objects by separating their respective subnets, and by implicitly retaining their mutual one-to-one and one-to-many relations through the notion of correlation set. In Figure~\ref{fig:clognet}, that would require to separate the subnets of orders, items, and trucks, relating them with two special one-to-many channels indicating that multiple items belong to the same order and loaded in the same truck.

A correlation set is established when one or multiple objects $\cval{o}_1,\ldots,\cval{o}_n$ are co-created, all being related to the same object $\cval{o}$ of a different type (cf.~the creation of multiple items for the same order in our running example). In Proclets, this correlation set is implicitly reconstructed by inspecting the concurrent histories of such different objects. Correlation sets are then used to formalise two sophisticated forms of synchronisation. In the \emph{equal} synchronisation, $\cval{o}$ flows through a transition $t_1$ while, simultaneously, \emph{all} objects $\cval{o}_1,\ldots,\cval{o}_n$ flow through another transition $t_2$. In the \emph{subset} synchronisation, the same happens but only requiring a subset of $\cval{o}_1,\ldots,\cval{o}_n$ to synchronise. 

Interestingly, \clognets can encode correlation sets and the subset synchronisation semantics. A correlation set is explicitly maintained in the net by imposing that the tokens carrying $\cval{o}_1,\ldots,\cval{o}_n$ also carry a reference to $\cval{o}$. This is what happens for items in our running example: they explicitly carry a reference to the order they belong to. Subset synchronisation is encoded via a properly crafted subnet. Intuitively, this subnet works as follows. First, a lock place is inserted in the \clognet so as to indicate when the net is operating in a normal mode or is instead executing a synchronisation phase. When the lock is taken, some objects in $\cval{o}_1,\ldots,\cval{o}_n$ are nondeterministically picked and moved through their transition $t_2$. The lock is then released, simultaneously moving $\cval{o}$ through its transition $t_1$. 
Thanks to this approach, a Proclet with subset synchronisation points can be encoded into a corresponding \clognet, providing for the first time a practical approach to verification. This does not carry over Proclets with equal synchronisation, which would allow us to capture, in our running example, sophisticated mechanisms like nsuring that when a truck moves to its destination, \emph{all} items contained therein are delivered.
Equal synchronisation can only be captured in \clognets by introducing a data-aware variant of wholeplace operation, which we aim to study in the future.

%% file: 5-conclusions.tex
\section{Conclusions}
We have brought forward an integrated model of processes and data founded on CPN that balances between modelling power and the possibility of carrying sophisticated forms of verification parameterised on read-only, immutable relational data. We have approached the problem of verification not only foundationally, but also showing a direct encoding into \mcmt, one of the most well-established model checkers for the verification of infinite-state dynamic systems. We have also shown that this model directly relates to some of the most sophisticate models studied in this spectrum, attempting at unifying their features in a single approach.
Given that \mcmt is based on Satisfiability Modulo Theories (SMT), our approach naturally lends itself to be extended with numerical data types and arithmetics. We also want to study the impact of introducing wholeplace operations, essential to capture the most sophisticated syhncronization semantics defined for Proclets \cite{Fahland19}.
At the same time, we are currently defining a benchmark for data-aware processes, systematically translating the artifact systems benchmark defined in \cite{verifas} into corresponding imperative data-aware formalisms, including \clognets.

%% file: 0-appendix.tex
\section{Execution Semantics}
\label{app:semantics}

\begin{figure}[h!]
\centering
\begin{tikzpicture}[->,>=stealth',auto,node distance=2.1cm,thick]
  \node[transition] (2) {$t$};
  \node[place,label=left:$p$] (1) [left of=2]{ \cname{a}};
  \node[draw=none,yshift=1.5cm] (lbl) [below of=1] {\datatype{String}};
  \path
    (1) edge[bend left=10] node[above] {$x$} (2)
    (2) edge[bend left=10] node[below] {$\nu$} (1);
\end{tikzpicture}
\caption{A simple marked \clognet}
\label{fig:net}
\end{figure}
Let $\markedcat{N}{m_0}{Cat}$ be a marked \clognet with catalog instance $Cat$. Then its execution semantics  is captured by transition system $\tsys{N}=(S, s_0, \trans)$, where:
\begin{compactitem}[$\bullet$]
\item $S$ is a possibly infinite set of markings over $N$;
\item $\trans\subseteq S\times \transitions \times S$ is a $\transitions$-labelled transition relation between pairs of markings;
\item $S$ and $\trans$ are defined by simultaneous induction as the smallest sets satisfying the following conditions:
\begin{inparaenum}[\it(i)]
\item $m_0\in S$;
\item given $m\in S$, for every transition $t\in\transitions$, binding $\sigma$ and marking $m'$ over $N$, if $\firecat{m}{t}{m'}{Cat}$, then $m'\in S$ and $m\overset{t}{\trans} m'$.
\end{inparaenum}
\end{compactitem}

In Figure~\ref{fig:net} we show a simple \clognet $N$ with an empty catalog, and with initial marking $m_0$ such that $m_0(p)=\set{\cname{a}}$. Its unique transition $t$ updates the content of place $p$ by replacing the current token with a token that carries a different (locally fresh) string value. Figure~\ref{fig:ts} depicts an infinite-state transition system representing the execution semantics of $N$. Even if not explicitly shown, each transition is labelled by $t$.

\definecolor{cobalt}{rgb}{0.0, 0.28, 0.67}
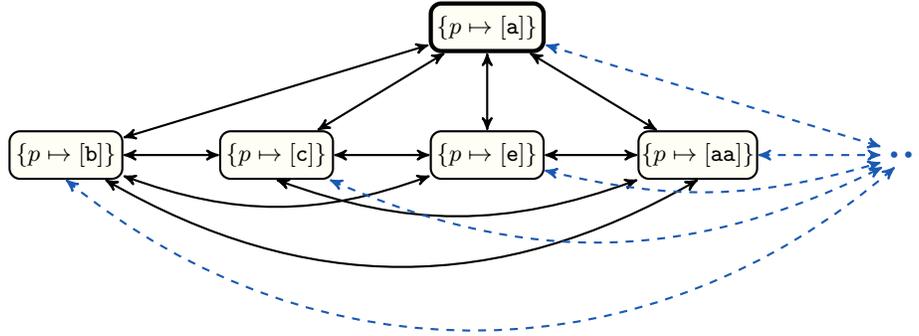
\begin{figure}
\centering
\begin{tikzpicture}[->,>=stealth',auto,x=2.8cm,y=1.7cm,thick]
 \node[state,ultra thick] (0) at (0,0) {$\left\{p \mapsto [\cname{a}]\right\}$};
\node[state] (1) at (-2,-1) {$\left\{p \mapsto [\cname{b}]\right\}$};
\node[state] (2) at (-1,-1) {$\left\{p \mapsto [\cname{c}]\right\}$};
\node[state] (3) at (0,-1) {$\left\{p \mapsto [\cname{e}]\right\}$};
\node[state] (4) at (1,-1) {$\left\{p \mapsto [\cname{aa}]\right\}$};
\node (5) at (2,-1)  {{\color{cobalt!90} \tiny $\bullet \bullet \bullet$}};
 \path
    (0) edge[<->] node[left] {} (1)
    (0) edge[<->] node[left] {} (2)
    (0) edge[<->] node[left] {} (3)
    (0) edge[<->] node[left] {} (4)
    (0) edge[<->,dashed,color=cobalt!90] node {} (5)
    (1) edge[<->] node {} (2)
    (1) edge[<->,bend right=20] node {} (3)
    (1) edge[<->,bend right=30] node {} (4.south)
    (1.south) edge[<->,bend right=40,dashed,color=cobalt!90] node {} (5)
    (2) edge[<->] node {} (3)
    (2.south) edge[<->,bend right=20] node {} (4.south west)
    (2) edge[<->,bend right=25,dashed,color=cobalt!90] node {} (5)
    (3) edge[<->] node {} (4)
    (3) edge[<->,bend right=15,dashed,color=cobalt!90] node {} (5)
    (4) edge[<->,dashed,color=cobalt!90] node {} (5)
;
\end{tikzpicture}
\caption{A transition system of a net from Figure~\ref{fig:ts}}
\label{fig:ts}
\end{figure}

%% file: 1-appendix.tex
\section{MCMT Encoding of the Initial Marking}
\label{app:initial-marking-mcmt}
A special \mcmt transition is used to inject the initial marking into the \mcmt array-based system. This \mcmt transition
populates the arrays representing places, initialised as all empty, with entries that correspond to the initial \clognet marking $m_0$. Please refer to Section~\ref{sec:translation} for the context of this encoding.

This \mcmt transition can be executed only if flag \texttt{init\_fl}, denoting whether the initial marking assignment has taken place, is \texttt{TRUE}.\footnote{In case of other statements, their guards $G$ should contain a conjunct \texttt{(= init\_fl FALSE)}.} It works as follows:
\begin{mc}
  :transition
  :var i1,..., iM
  :var j
  :guard (=init_fl TRUE) 
  :numcases NCm0
  ...
  :case (= j $i$)
  :val $v_{1,i}$
  ...
  :val $v_{k,i}$
  ...
  :val FALSE
  ...
\end{mc}

 Note that the flag should be previously declared using the \mcmt statement \texttt{:global  init\_fl BOOLE}. Same holds for the boolean constants \texttt{TRUE} and \texttt{FALSE}: they are declared using the respective statements \texttt{:smt (define TRUE ::BOOLE)} and \texttt{:smt (define FALSE ::BOOLE)}. 
Then, the transition runs through \texttt{NCm0} cases. 
All the cases go over the indexes \texttt{i1},\ldots, \texttt{iM} that correspond to tokens that have to be added to places.
More specifically, for every place $p\in\places$ such that $m_0(p)\neq\mult\emptyset$ and $|\coloring{p}|=k$, we add an $i$-th token to it by putting constant $v_{r,i}\in C$ in $i$-th place of every $r$-th component array of $p$. Moreover, every case has to update \texttt{init\_fl}, changing its value to   \texttt{FALSE}.

%% file: main.bbl
\begin{thebibliography}{10}
\providecommand{\url}[1]{\texttt{#1}}
\providecommand{\urlprefix}{URL }
\providecommand{\doi}[1]{https://doi.org/#1}

\bibitem{Aalst19}
van~der Aalst, W.M.P.: Object-centric process mining: Dealing with divergence
  and convergence in event data. In: Proc.\ of SEFM. LNCS, vol. 11724, pp.
  3--25. Springer (2019)

\bibitem{AGPR12}
Alberti, F., Ghilardi, S., Pagani, E., Ranise, S., Rossi, G.P.: Universal
  guards, relativization of quantifiers, and failure models in model checking
  modulo theories. J. Satisf. Boolean Model. Comput.  \textbf{8}(1/2),  29--61
  (2012)

\bibitem{AKMA19}
Artale, A., Kovtunova, A., Montali, M., van~der Aalst, W.M.P.: Modeling and
  reasoning over declarative data-aware processes with object-centric
  behavioral constraints. In: Proc.\ of BPM. LNCS, vol. 11675, pp. 139--156.
  Springer (2019)

\bibitem{BaHW17}
Batoulis, K., Haarmann, S., Weske, M.: Various notions of soundness for
  decision-aware business processes. In: Proc.\ of ER. LNCS, vol. 10650, pp.
  403--418. Springer (2017)

\bibitem{CGGMR18}
Calvanese, D., Ghilardi, S., Gianola, A., Montali, M., Rivkin, A.: Verification
  of data-aware processes via array-based systems (extended version). Technical
  Report arXiv:1806.11459, arXiv.org (2018)

\bibitem{BPM19}
Calvanese, D., Ghilardi, S., Gianola, A., Montali, M., Rivkin, A.: Formal
  modeling and {SMT}-based parameterized verification of data-aware {BPMN}. In:
  Proc.\ of {BPM}. LNCS, vol. 11675. Springer (2019)

\bibitem{CGGMR19-techrep-dab}
Calvanese, D., Ghilardi, S., Gianola, A., Montali, M., Rivkin, A.: Formal
  modeling and {SMT}-based parameterized verification of data-aware {BPMN}
  (extended version). {Technical Report arXiv:1906.07811}, arXiv.org (2019)

\bibitem{CGGMR19}
Calvanese, D., Ghilardi, S., Gianola, A., Montali, M., Rivkin, A.: From model
  completeness to verification of data aware processes. In: Description Logic,
  Theory Combination, and All That. LNCS, vol. 11560. Springer (2019)

\bibitem{MSCS20}
Calvanese, D., Ghilardi, S., Gianola, A., Montali, M., Rivkin, A.: {SMT}-based
  verification of data-aware processes: a model-theoretic approach.
  Mathematical Structures in Computer Science  \textbf{30}(3),  271--313
  (2020). \doi{10.1017/S0960129520000067}

\bibitem{CaDM13}
Calvanese, D., De~Giacomo, G., Montali, M.: Foundations of data aware process
  analysis: {A} database theory perspective. In: Proc.\ of PODS. ACM (2013)

\bibitem{DDG17}
De~Masellis, R., Di~Francescomarino, C., Ghidini, C., Montali, M., Tessaris,
  S.: Add data into business process verification: Bridging the gap between
  theory and practice. In: Singh, S.P., Markovitch, S. (eds.) Proc.\ of AAAI.
  pp. 1091--1099 (2017)

\bibitem{DeLV16}
Deutsch, A., Li, Y., Vianu, V.: Verification of hierarchical artifact systems.
  In: Proc.\ of PODS. pp. 179--194. ACM (2016)

\bibitem{Dum11}
Dumas, M.: On the convergence of data and process engineering. In: Proc.\ of
  ABDIS. LNCS, vol.~6909, pp. 19--26. Springer (2011)

\bibitem{Fahland19}
Fahland, D.: Describing behavior of processes with many-to-many interactions.
  In: Proc. of Petri nets 2019. pp. 3--24. Springer (2019)

\bibitem{lmcs}
Ghilardi, S., Ranise, S.: Backward reachability of array-based systems by {SMT}
  solving: Termination and invariant synthesis  \textbf{6}(4) (2010)

\bibitem{HSVW09}
van Hee, K.M., Sidorova, N., Voorhoeve, M., van~der Werf, J.M.E.M.: Generation
  of database transactions with petri nets. Fundam. Inform.  \textbf{93}(1-3),
  171--184 (2009)

\bibitem{Hull08}
Hull, R.: Artifact-centric business process models: {B}rief survey of research
  results and challenges. In: Proc.\ of ODBASE. pp. 1152--1163 (2008)

\bibitem{KuWR11}
K{\"u}nzle, V., Weber, B., Reichert, M.: Object-aware business processes:
  Fundamental requirements and their support in existing approaches. Int.\ J.\
  of Information System Modeling and Design  \textbf{2}(2),  19--46 (2011)

\bibitem{verifas}
Li, Y., Deutsch, A., Vianu, V.: {VERIFAS:} {A} practical verifier for artifact
  systems  \textbf{11}(3),  283--296 (2017)

\bibitem{MPFW13}
Meyer, A., Pufahl, L., Fahland, D., Weske, M.: Modeling and enacting complex
  data dependencies in business processes. In: Proc.\ BPM. LNCS, vol.~8094, pp.
  171--186. Springer (2013)

\bibitem{MonR17}
Montali, M., Rivkin, A.: {DB-Nets}: on the marriage of colored {Petri} {Nets}
  and relational databases. Trans. of Petri Nets and other Models of
  Concurrency  \textbf{28}(4) (2017)

\bibitem{MonR19}
Montali, M., Rivkin, A.: From db-nets to coloured petri nets with priorities.
  In: Proc.\ of PN. pp. 449--469 (2019)

\bibitem{PWOB19}
Polyvyanyy, A., van~der Werf, J.M.E.M., Overbeek, S., Brouwers, R.: Information
  systems modeling: Language, verification, and tool support. In: Proc. of
  CAiSE 2019. LNCS, vol. 11483, pp. 194--212. Springer (2019)

\bibitem{Reic12}
Reichert, M.: Process and data: {T}wo sides of the same coin? In: Proc.\ of
  OTM. pp. 2--19 (2012)

\bibitem{RVFE11}
Rosa-Velardo, F., de~Frutos-Escrig, D.: Decidability and complexity of petri
  nets with unordered data. Theor. Comput. Sci.  \textbf{412}(34),  4439--4451
  (2011)

\end{thebibliography}
